\documentclass[10pt,twocolumn,letterpaper]{article}

\usepackage{wacv}
\usepackage{times}
\usepackage{epsfig}
\usepackage{graphicx}
\usepackage{amsmath}
\usepackage{amssymb}
\usepackage{booktabs}
\usepackage{float}
\usepackage{ragged2e}
\usepackage{enumitem}
\usepackage{authblk}
\usepackage[caption=false]{subfig}
\usepackage[accsupp]{axessibility}
\usepackage{pifont}% http://ctan.org/pkg/pifont
\newcommand{\cmark}{\ding{51}}%
\newcommand{\xmark}{\ding{55}}%
\def\wacvPaperID{855} % Enter the WACV Paper ID here

\wacvfinalcopy % *** Uncomment this line for the final submission

\ifwacvfinal
\usepackage[breaklinks=true,bookmarks=false]{hyperref}
\else
\usepackage[pagebackref=true,breaklinks=true,colorlinks,bookmarks=false]{hyperref}
\fi
\begin{document}

%%%%%%%%% TITLE
\title{MovieCLIP: Visual Scene Recognition in Movies}

% \author{Digbalay Bose\\
% University of Southern California\\
% Los Angeles, CA\\
% {\tt\small dbose@usc.edu}
\author[1]{Digbalay Bose}
\author[1]{Rajat Hebbar}
\author[2]{Krishna Somandepalli}
\author[1]{Haoyang Zhang}
\author[2]{Yin Cui}
\author[2]{Kree Cole-McLaughlin}
\author[2]{Huisheng Wang}
\author[1]{Shrikanth Narayanan}

% To save space, use either the email address or home page, not bothhttps://www.overleaf.com/project/62bfbf839e316bc0d4ca131b
% \and
\makeatletter \renewcommand\AB@affilsepx{\hspace{3mm} \protect\Affilfont} \makeatother
\affil[1]{University of Southern California, Los Angeles, CA}
\affil[2]{Google}
\affil[1]{\tt {{\small\{dbose@,rajatheb@,zhangh21@,shri@ee.\}usc.edu}}}
\affil[2]{\tt {{\small\{ksoman@,yincui@,kree@,huishengw@\}google.com}}}
%\affil[1]{University of Southern California, Los Angeles, CA}
\maketitle
\thispagestyle{empty}

%%%%%%%%% ABSTRACT
\begin{abstract}
   Longform media such as movies have complex narrative structures, with events spanning a rich variety of ambient visual scenes. 
   %A key element therein is the identification of visual scenes of people in potentially diverse contexts. 
   Domain specific challenges associated with visual scenes in movies include transitions, person coverage, and a wide array of real-life and fictional scenarios. Existing visual scene datasets in movies have limited taxonomies and don’t consider the visual scene transition within movie clips. In this work, we address the problem of visual scene recognition in movies by first automatically curating a new and extensive movie-centric taxonomy of 179 scene labels derived from movie scripts and auxiliary web-based video datasets. Instead of manual annotations which can be expensive, we use CLIP to weakly label 1.12 million shots from 32K movie clips based on our proposed taxonomy. We provide baseline visual models trained on the weakly labeled dataset called \textbf{MovieCLIP} and evaluate them on an independent dataset verified by human raters. We show that leveraging features from models pretrained on MovieCLIP benefits downstream tasks such as multi-label scene and genre classification of web videos and movie trailers.
\end{abstract}
%%%%%%%%% BODY TEXT
\section{Introduction}
Media, in its diverse forms and modalities, is used to create and share narratives across domains including movies, television shows, advertisements, games, news, and user generated social stories. Movies represent a major form of media content,  with box office revenues estimated at \$4.48 billion across 329 movies released in 2021 \cite{statisticsmovie}, with a global reach and societal influence. 
\begin{figure}[!h]
 \centering
  \includegraphics[width=\columnwidth]{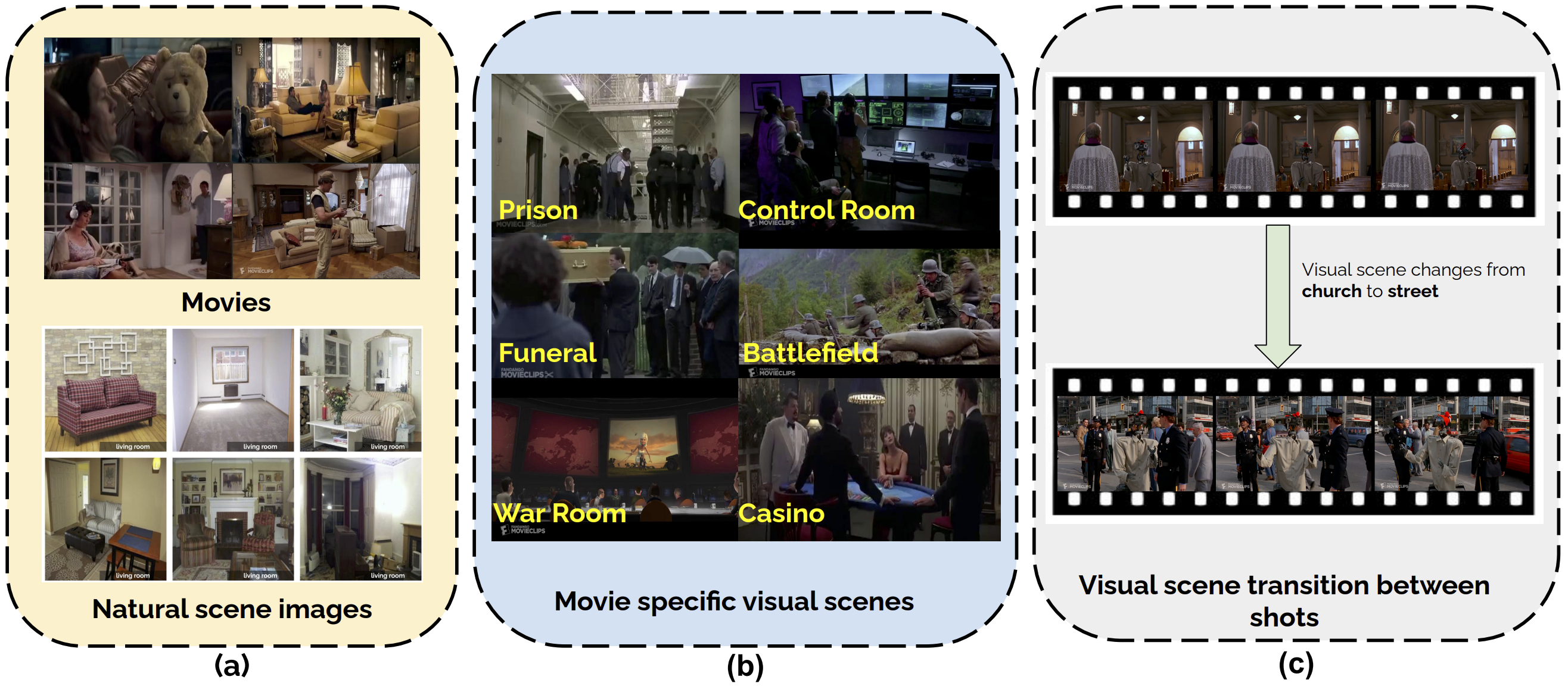}
  \caption{Overview diagram highlighting the challenges associated with visual scene recognition in movies (a) Domain mismatch between Natural scene images,(Source: \url{http://places2.csail.mit.edu/explore.html}) vs frames from Movies for \textbf{living room} (b) Movie centric visual scene classes like prison, control room etc that are absent from existing taxonomies (c) Change in visual scene between shots in the same movie clip.}
  \label{Intro figure}
  \label{fig:teaser}
\end{figure}
The computational analysis of media content \cite{CMI} especially movies presents unique challenges due to their long-form narrative structures with character interactions often spanning diverse visual scenes and contexts. 
In cinematic terms, \textit{mis-en-scene} \cite{bordwellthomson} refers to how the different elements of a film are depicted and arranged in front of camera. Key components of \textit{mis-en-scene} include the actors with their different styles, visual scenes where the interactions take place, set design including lighting and camera placement and the accompanying costumes and makeup of the artists. The visual scene is considered a crucial component since it sets the mood and provides a background for the various actions performed by the actors in the scene. Visual scenes in movies are often tied to social settings like weddings, birthday parties and workplace gatherings that provide information about character interactions.
\par 
Accurate recognition of visual scenes can help in uncovering the bias involved in portrayal of under-represented characters vis-a-vis different scenes e.g., fewer women shown in office as compared to kitchen. For content tagging tasks like genre classification, visual scenes provide context information like battlefield portrayals in action/adventure movies, space-shuttle in sci-fi movies or courtrooms in dramas.
However, there are certain inherent challenges in visual scene recognition for movies that needs to be addressed, as shown in Fig.~\ref{Intro figure}:
\\
\textbf{Domain mismatch - scene images vs movie frames:} Visual scenes depicted in movies are distinct compared to natural scenes due to increased focus on actors, multiple activities and viewpoint variations like extreme closeup, wide angle shots etc. An example is shown in Fig.~\ref{Intro figure} (a) for images from Places2 dataset \cite{zhou2017places} and movie frames from Condensed Movies dataset \cite{bain2020condensed}.
\\
\textbf{Lack of completeness in scene taxonomy:} Movies depict both real life and fictional scenarios that span a wide variety of visual scenes. As shown in Fig.~\ref{Intro figure}(b), certain movie centric visual scene classes like \textit{battlefield}, \textit{control room}, \textit{prison}, \textit{war room}, \textit{funeral}, \textit{casino} are absent from existing public scene taxonomies associated with natural scene image and video datasets.\\
\textbf{Lack of shot specific visual scene annotations:}
Existing datasets like Condensed Movies \cite{bain2020condensed} and VidSitu \cite{Sadhu_2021_CVPR} provide a {\em single} visual scene label for the entire movie clip (around 2 minutes long), obtained through descriptions provided as part of YouTube channel Fandango Movie clips \footnote{https://www.youtube.com/channel/UC3gNmTGu-TTbFPpfSs5kNkg}. In Fig.~\ref{Intro figure} (c), the provided description: \textit{Johnny Five (Tim Blaney) searches for his humanity in the \textbf{streets} of New York.} mentions only the visual scene \textbf{street}, while the initial set of events takes place inside \textbf{church}. Instead of considering a single scene label for the entire movie clip, shot level visual scene annotation can help in tracking the scene change from \textbf{church} to \textbf{street}.
\par
In our work, we consider shots within a given movie clip as the fundamental units for visual scene analysis since shots consist of consecutive set of frames related to the same content, whose starting and ending points are triggered by recording using a single camera \cite{SBD}. Our contributions are as follows:
\begin{itemize}[leftmargin=*]
\addtolength\itemsep{-2mm}
\item \textbf{Movie-centric scene taxonomy:} We develop a movie-centric scene taxonomy by leveraging scene headers (sluglines) from movie scripts and existing video datasets with scene labels like HVU\cite{diba_large_2020}. 
%using a human-in-the loop strategy for taxonomy refinement.
\item \textbf{Automatic shot tagging:} We utilize our generated scene taxonomy to automatically tag around 1.12M shots from 32K movie clips using CLIP \cite{CLIP} based on a frame-wise aggregation scheme.  
\item \textbf{Multi-label scene classification:} We develop multi-label scene classification baselines using the shot-level tagged dataset called MovieCLIP and evaluate them on an independent shot level dataset curated by human experts. The dataset and associated codebase can be accessed at \url{https://sail.usc.edu/mica/MovieCLIP/}
%\footnote{The dataset will be publicly released upon acceptance} a
\item \textbf{Downstream tasks:} We further extract feature representations from the baseline models pretrained on MovieCLIP and explore their applicability in diverse downstream tasks of multi-label scene and movie genre classification from web videos \cite{diba_large_2020} and trailers \cite{2019Moviescope},  respectively. 
\end{itemize}
\section{Related work}
\textbf{Image datasets for visual scene recognition:}
Image datasets for scene classification like MIT Indoor67 \cite{IndoorScenes} relied on categorizing a finite set of (67) indoor scene classes. A broad categorization into indoor, outdoor (natural) and outdoor (man-made) groups for 130K images across 397 subcategories was introduced by the SUN dataset \cite{xiao_sun_2016}. For large scale scene recognition, the Places dataset \cite{zhou2017places} was developed with 434 scene labels spanning 10 million images. The scene taxonomy considered in Places dataset was derived from the SUN dataset, followed by careful merging of similar pairs. It should be noted that the curation of large scale visual scene datasets like Places relied on crowd-sourced manual annotations over multiple rounds.\\
\textbf{Video datasets for visual scene recognition:} While there has been considerable progress in terms of action recognition capabilities from videos due to introduction of large scale datasets like Kinetics \cite{kinetics400}, ActivityNet \cite{caba2015activitynet}, AVA \cite{gu2018ava}, Something-Something \cite{Something-SomethingV2}, only few large scale datasets like HVU \cite{diba_large_2020} and Scenes, Objects and Actions (SOA) \cite{SOA} have focused on scene categorization with actions and associated objects. SOA was introduced as a multi-task multi-label dataset of social-media videos across 49 scenes with objects and actions but the taxonomy curation involves free-form tagging by human annotators followed by automatic cleanup. %However, SOA being publicly not available, its wide scale adoption by the community is limited. 
HVU \cite{diba_large_2020}, a recently released public dataset of web videos with 248 scene labels, relied on initial tag generation based on cloud APIs followed by human verification.\\
\textbf{Movie-centric visual scene recognition:} In the domain of scene recognition from movies, Hollywood scenes \cite{marszalek09} was first introduced with 10 scene classes extracted from headers in movie scripts across 3669 movie clips. A socially grounded approach was explored in Moviegraphs \cite{moviegraphs} with emphasis on the underlying interactions (relationships/situations) along with spatio-temporal localizations and associated visual scenes (59 classes).
\begin{table*}[h!]
\centering
\resizebox{0.7\textwidth}{!}{
\begin{tabular}{|c|c|c|c|c|c|c|}
\hline
\textbf{Dataset} & \textbf{Domain} & \textbf{\#classes} & \textbf{\#samples} & \textbf{Annotation} & \textbf{Unit} & \textbf{AV} \\ \hline
Scene 15 \cite{BayesianFeiFeiLi}   & Natural & 15  & $\sim$6k   & Manual & Image  & \cmark     \\ \hline
MITIndoor67  \cite{IndoorScenes} & Natural     & 67   & 15620   & Manual  & Image &\cmark    \\ \hline
SUN397  \cite{xiao_sun_2016}   & Natural    & 397   & 130,519  & Manual  & Image &\cmark    \\ \hline
Places  \cite{zhou2017places}    & Natural   & 434   & 10m  & Manual  & Image & \cmark     \\ \hline
Hollywood Scenes \cite{marszalek09}   &  Movies    & 10    & 3669   & Automatic  & Video clip (36.1s)  & \cmark  \\ \hline
Moviegraphs  \cite{moviegraphs}    & Movies   & 59     & 7637   & Manual  & Video clip (44.28 s)  & \xmark    \\ \hline
SOA \cite{SOA} &  Web-Videos   & 49   & 562K   & Semi-automatic  & Video clip (10 s)  &   \xmark      \\ \hline
Movienet  \cite{huang2020movienet}  & Movies   & 90    & 42K  & Manual  & Scene segment (2 min)  & \xmark           \\ \hline
HVU    \cite{diba_large_2020}    & Web-Videos    & 248    & 251k  & Semi-automatic &  Video clip (10 s) & \cmark    \\ \hline
Condensed Movies \cite{bain2020condensed}   & Movies    & NA    & 33k   & Automatic  & Video clip ( 2 min) & \cmark    \\ \hline
VidSitu  \cite{Sadhu_2021_CVPR}    & Movies  & $\sim$50    & 14k   & Manual  & Video  clip (10 s) & \cmark     \\ \hline
LVU  \cite{lvu2021}    & Movies    & 6    & 723   & Automatic &  Video clip ($1 \sim 3$ min) & \cmark       \\ \hline
\textbf{MovieCLIP}  & \textbf{Movies}  & \textbf{179}  & \textbf{1.12m}& \textbf{Automatic} & \textbf{Shot (3.54s) }  & \textbf{\cmark }    \\ \hline
\end{tabular}
}
\vspace{5mm}
\caption{Comparison of MovieCLIP with other available image and video datasets with visual scene classes.  \textbf{Natural}: Images of natural scenes.\textbf{Web-Videos}: videos obtained from internet sources like YouTube. \textbf{AV}: whether publicly available or not. Avg or duration span of video data sources are provided with the respective units. \textbf{NA}: Number of scene classes explicitly not mentioned with the dataset.}
\label{Overview}
\end{table*}
For holistic movie understanding tasks, the Movienet dataset\cite{huang2020movienet} was introduced with the largest movie-centric scene taxonomy consisting of 90 place (visual scene) tags with segment wise human annotations of entire movies. Instead of entire movie data, short movie clips sourced from YouTube channel of Fandango Movie clips were used for text-video retrieval in Condensed movies dataset \cite{bain2020condensed}, visual semantic role labeling \cite{Sadhu_2021_CVPR} and pretraining object-centric transformers \cite{transformers} for long-term video understanding in LVU dataset \cite{lvu2021}. While there is no explicit visual scene labeling, the raw descriptions available on Youtube with the movie clips have mentions of certain visual scene classes.
% Instead of entire movie data, short movie clips sourced from YouTube channel of Fandango Movie clips were provided as part of Condensed movies dataset \cite{bain2020condensed}. While there is no explicit visual scene labeling in Condensed Movies, the raw descriptions provided as part of movie clips have mentions of the visual scene classes. 
% The short movie snippets available as part of Fandango Movie clips were further used for tasks like visual semantic role labeling \cite{Sadhu_2021_CVPR} and pretraining object-centric transformers \cite{transformers} for long-term video understanding in LVU dataset \cite{lvu2021}. 
\\
MovieCLIP, our curated dataset, is built on top of movie clips available as a part of Condensed Movies dataset \cite{bain2020condensed}. A comparative overview of MovieCLIP and other image and video datasets with visual scene labels is shown in Table~\ref{Overview}.
In comparison with previous video-centric works, our taxonomy generation relies on domain-centric data sources like movie scripts and auxiliary world knowledge from web-video based sources like HVU with minimal human-in-the loop supervision for taxonomy refinement.
\\
\textbf{Knowledge transfer from pretrained vision language models:}
Vision language(V-L) based pretraining methods involve learning transferable visual representations based on various pretext tasks associated with image and text pairs. Examples of pretext tasks in V-L domain include prediction of masked words in captions based on visual cues in ICMLM \cite{sariyildiz2020icmlm}, pretraining image encoders based on bicaptioning objective in VirTex \cite{desai2021virtex} and contrastive alignment of image-caption pairs in CLIP \cite{CLIP}. Leveraging features from CLIP's visual and text encoders have improved existing vision-language tasks \cite{Shen2021HowMC} and enabled open-vocabulary object detection \cite{Gu2021OpenvocabularyOD}, and language driven semantic segmentation \cite{Li2022LanguagedrivenSS}.
% An example of pretext task includes prediction of masked words in captions based on visual cues in ICMLM \cite{sariyildiz2020icmlm} and further finetuning for downstream tasks like scene and object recognition. Further usage of pretrained image encoders based on a bicaptioning pretext task was proposed in VirTex \cite{desai2021virtex}, where downstream evaluation included tasks like instance segmentation and object detection. While the above mentioned methods used prediction of caption tokens based on visual inputs, a contrastive pretraining approach based on alignment of images-captions was introduced by CLIP \cite{CLIP}. 
In our work, we use the pretrained visual and text encoders of CLIP and utilize it as a noisy annotator by tagging movie shots based on our curated visual scene taxonomy.
\section{Taxonomy curation for movie scenes}
In this section we outline the process involved in curating a visual scene taxonomy based on the domain information present in movie scripts and the pre-existing scene information present in auxiliary video datasets.
\subsection{Sources of visual scene information}
Movie scripts have been used as external sources for describing and annotating videos through script and subtitle alignment methods in \cite{10.1007/978-3-540-88693-8_12}, \cite{Everingham2006HelloMN}, \cite{Laptevactionscvpr}, \cite{Rohrbach2015ADF}. Movie scripts contain \textit{sluglines} that provide information about visual scene, time of the day, and whether the action takes place in indoor or outdoor settings.
Sample example of a slugline with visual scene as river is : \uppercase{EXT. GOTHAM \textbf{RIVER} - DAY}.
%Sample sluglines are shown in Fig \ref{script labels}, with the visual scenes marked in red. 
We parse 156k sluglines from a in-house set of 1434 movie scripts. For each slugline, we automatically extract entities after the ``EXT.''(exterior) or ``INT.''(interior) tags like \textit{Hospital room}, \textit{River}, \textit{War room} etc.
Using this procedure, we extract 173 unique visual scene labels. Since our taxonomy generation process is motivated by visual scenes in movies with sluglines from scripts as seed sources along with auxiliary sources.
Since the set of labels from movie scripts is not exhaustive, we also consider auxiliary sources especially web video datasets with visual scene labels like HVU \cite{diba_large_2020}. We consider HVU as source of additional labels since the taxonomy (248 visual scene classes) is semi-automatically curated for short-trimmed videos, having similar nature to movie shots. We don’t consider Places2 \cite{zhou2017places} dataset since the taxonomy is primarily curated for natural
scenes, which are distinct from movie-centric visual scenes.
%
% \begin{figure}[h!]
% \centering
% %\subfloat[]{%
% \includegraphics[width=0.7\columnwidth]{Figures/Movie_script_organization.png}%
% \label{script labels}
% \caption{(a) Sample sluglines from movie script of ``The Dark Knight Rises(2012)'' showing the variety in visual scenes (surrounded by red box) }
% \label{script labels}
% \end{figure}

\subsection{Visual scene taxonomy curation}
In order to develop a comprehensive taxonomy of visual scenes in movies, we develop an automatic way of merging taxonomies from movie sluglines and the auxiliary dataset i.e., HVU with minimal human-in-the-loop for post-processing. The broad steps involved in the taxonomy generation are listed as below:
\par
\textbf{Label space preprocessing:} For simplicity, we consider the set of unique labels from movie sluglines (MS) to be denoted as $L_{MS}$ and its cardinality as $N_{MS}$. Similarly we denote $N_{HVU}$ to be the cardinality of the set $L_{HVU}$, i.e., the set of unique labels from the HVU dataset. For our case, we have $N_{MS}=173$ and $N_{HVU}=248$. We extract the intersecting set of labels between $L_{MS}$ and $L_{HVU}$, denoted by the set $L_{com}$ . The number of common labels between HVU and movie slugline based taxonomy is $N_{com}=68$. We remove the common set of labels $L_{com}$ from both the label spaces of movie sluglines and HVU. This gives us a non-intersecting set of labels in movie sluglines and HVU denoted by $L_{MS} \setminus L_{Com}$  and $L_{HVU} \setminus L_{Com}$ respectively.  We combine the sets of labels i.e. $L_{MS} \setminus L_{Com}$ and $L_{HVU} \setminus L_{Com}$ to obtain a larger set of labels called $L_{NC}$, where NC refers to not common. 
\begin{equation}
    L_{NC}=(L_{MS} \setminus L_{Com}) \cup (L_{HVU} \setminus L_{Com})
\end{equation}
\par 
\textbf{Merging with common label space:}
In this step, we find labels in $L_{NC}$ that are semantically close to the labels in $L_{com}$. We extract dense 384D label representations using the MiniLM-L6-v2 sentence transformers model \cite{reimers-2019-sentence-bert} for labels in $L_{NC}$ and $L_{com}$. For each label in $L_{NC}$ we compute cosine similarities with set of labels in $L_{com}$ based on label representations. We merge those labels from $L_{NC}$ with the similar labels in $L_{com}$, whose top-1 cosine similarity values are greater than 0.6. We update the set of labels $L_{NC}$ to $L_{N}$ by removing the merged labels.
Examples of such merging with respective cosine similarities and sources are as follows:
\begin{flushleft}
\begin{itemize}[leftmargin=*]
\addtolength\itemsep{-2mm}
    \item $\textbf{dune} \{L_{NC}\} \rightarrow{} \textbf{desert} \{L_{com}\} (0.66)$ 
    \item $\textbf{tennis} \ camp \{L_{NC}\} \rightarrow{} \textbf{tennis \ court} \{L_{com}\}  (0.62)$
    \item $\textbf{restroom} \{L_{NC}\} \rightarrow{} \textbf{bathroom} \{L_{com}\} (0.80)$
    \item $\textbf{rural \ area} \{L_{NC}\} \rightarrow{} \textbf{village} \{L_{com}\} (0.64)$
    \item $\textbf{boardwalk} \{L_{NC}\} \rightarrow{} \textbf{walkway} \{L_{com}\} (0.67)$
    \item $\textbf{television \ room} \{L_{NC}\} \rightarrow{} \textbf{living \ room} \{L_{com}\} (0.67)$
    \item $\textbf{glacial \ lake} \{L_{NC}\} \rightarrow{} \textbf{lake} \{L_{com}\} (0.73)$
\end{itemize}
\end{flushleft}

\begin{figure}[h!]
\centering
%\subfloat[]{%
\includegraphics[width=0.7\columnwidth]{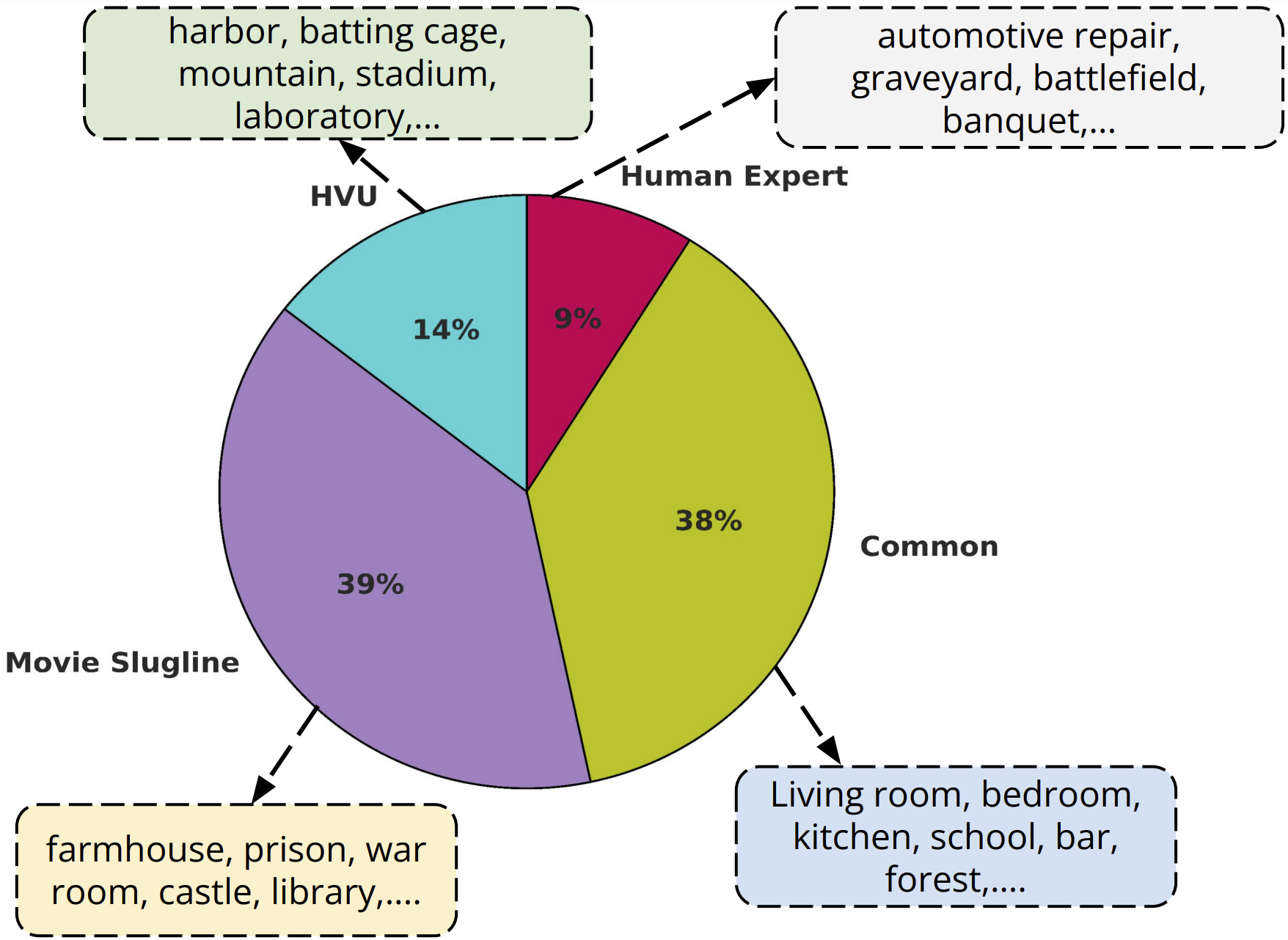}%
\caption{
The share of different sources (\textbf{HVU}, \textbf{Movie Sluglines}, \textbf{Common labels and Human expert}) in curating the label taxonomy. Example labels from different sources are shown in boxes with the pie chart.
}
\label{label distribution}
\end{figure}

\textbf{Human-in-the-loop taxonomy refinement:} 
% A human expert inspects the remaining labels in $L_{N}$ and bins them into relevant and irrelevant groups. Irrelevant group includes generic scene labels like \textit{body of water}, \textit{coastal and oceanic landforms}, \textit{horizon}, \textit{landscape}, \textit{underground} etc or highly specific scene labels with low probability of occurrence in movie scripts like \textit{coral reef}, \textit{white house}, \textit{piste}, \textit{badlands} etc. For relevant labels, we use the label representations from previous step and compute pairwise cosine similarities . 
A human expert inspects the labels in $L_{N}$ and removes both generic scene labels such as \textit{body of water}, \textit{coastal and oceanic landforms}, \textit{horizon}, \textit{landscape}, \textit{underground} as well as highly specific scenes such as \textit{coral reef}, \textit{white house}, \textit{piste}, \textit{badlands} etc. We use label representations from the previous step to exploit semantic similarity between the labels remaining in $L_{N}$.
For each relevant label in $L_{N}$, a threshold of 0.7 on top-1 similarity score is used to filter out similar labels. While comparing two similar labels, the human expert relies on wiki definitions to merge the more specific label into the generic one. For example, by definition \textit{bazaar} is a special form of market selling local items (as per wiki) and therefore merged with the \textit{market}. Other examples include:
\begin{itemize}[leftmargin=*]
    \addtolength\itemsep{-2.5mm}
    \item $\{stream, river bed, creek, river\} \rightarrow{} river$ 
    \item $\{hill, mountain, mountain \ pass, mountain \ range\} \\ \rightarrow{} mountain$
    \item $\{road, road \ highway, lane \} \rightarrow{} road$
    \item $\{port, marina \ dock, harbor \} \rightarrow{} harbor$
\end{itemize}
This results in a set of labels called $L_{merge}$ from $L_{N}$. Further, The human expert is exposed to randomly sampled 1000 shots from movie clips in Condensed Movies \cite{bain2020condensed} and reviews the current set of labels in the set $L_{merge} \cup L_{com}$. Based on the video content, a set of scene labels ($L_{human}$) that are missing from the current set is added by human expert. Thus the final set of 179 visual scene labels is obtained as follows:
\begin{equation}
     L_{final}=L_{com}\cup L_{merge} \cup L_{human}
\end{equation}
\textbf{Label source distribution:} As shown in Fig. \ref{label distribution}, the largest share is from movie sluglines (39\%).Only 9\% of the total labels is provided through feedback from human expert. Instead of manually binning classes into broad categories like indoor, outdoor or man-made, we discover groupings among classes through Affinity propagation clustering \cite{brendanfrey}  (based on label representations from sentence transformers). Certain clusters of visual scene labels are listed below, where all the sport locations, water bodies and performing arts locations are grouped together.
\begin{itemize}[leftmargin=*]
\addtolength\itemsep{-2.5mm}
    \item \textbf{sport locations:} Basketball court, Race track, Tennis court, Batting cage, Golf course
    \item \textbf{water bodies:} River, Pool, Waterfall, Hot spring, Pond, Swamp, Lake
    \item \textbf{performing arts locations:} Stage, Conference Room, Theater, Auditorium, Ballroom
    \item \textbf{Natural landforms:} Mountain, Desert, Valley
\end{itemize}
A detailed list of the clusters discovered among the visual scene classes is shown in Supplementary.
\section{MovieCLIP dataset}
\begin{figure*}[h]
\centering
\includegraphics[width=0.7\textwidth]{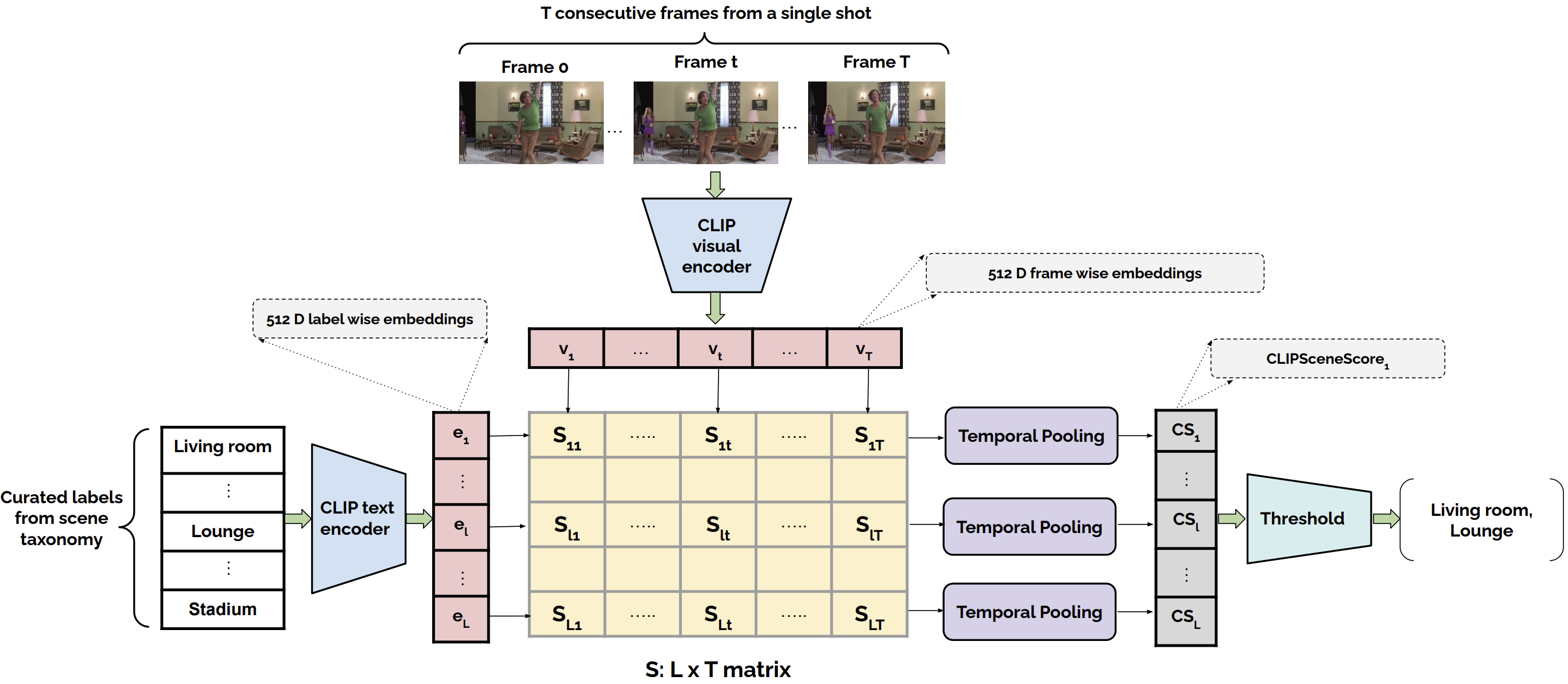} %0.8\textwidth
\caption{Overview schematic of the prompt based visual scene labeling of movie shot using CLIP's visual and text encoders. $S$ is the similarity matrix where entry $S_{lt}$ refers to similarity values between textual embedding $e_{l}$ and visual embedding $v_{t}$.}
\label{clip tagging}
\end{figure*}
We use the curated taxonomy described in Section 3 to develop a labeled dataset of movie shots called MovieCLIP. 
%The dataset and associated codebase will be made public upon acceptance. 
We outline the process of shot detection and automatic labeling using CLIP \cite{CLIP} in the following sections:
\subsection{Shot detection from movie clips}
Since a shot represents a continuous set of frames with minimal change in the visual scene, we consider movie shots for our subsequent analysis. Associating visual scene labels to shots can help in identifying cases where visual scene recognition is difficult like closeup or extreme closeup scenarios, even within the same movie scene. A singular movie scene involves changes in camera viewpoints across consecutive shots, thus making it difficult for CLIP to associate labels with high confidence for the entire movie scene.
For shot detection, we use PySceneDetect\footnote{https://pyscenedetect.readthedocs.io/en/latest/} to segment the movie clips in Condensed Movies with default parameters and content-aware detection mode. The overall statistics of the shot extraction process is shown in Table \ref{shot stats table}:
\begin{table}[h!]
\centering 
\resizebox{\columnwidth}{!}{
\begin{tabular}{|l|c|l|l|l|l|l|}
\hline
\textbf{\# movies} & \textbf{Years} & \textbf{\# clips}    & \textbf{\# shots} & \textbf{Avg shots/clip} & \textbf{Avg Duration}       \\ \hline
\multicolumn{1}{|c|}{3574} & 1930-2019      & \multicolumn{1}{c|}{32484} & 1124638           & \multicolumn{1}{c|}{34.66}&3.54s \\ \hline
\end{tabular}
}
\vspace{1mm}
\caption{Statistics of movie shots in MovieCLIP dataset.}
\label{shot stats table}
\end{table}
\subsection{CLIP based visual scene labeling of movie shots}
In this section, we describe how CLIP \cite{CLIP} can be used to associate visual scene labels with individual movie shots. 
Since CLIP has been trained in a contrastive manner for alignment, it
can be used to develop zero-shot classifiers for different tasks including scene recognition (\cite{xiao_sun_2016}), fine grained classification (\cite{parkhi12a}, \cite{bossard14}, \cite{KrauseStarkDengFei-Fei_3DRR2013}), facial emotion recognition (\cite{BarsoumICMI2016}), object (\cite{deng2009imagenet},\cite{FeiFei2004LearningGV}) and action classification (\cite{Kinetics700},\cite{ucf101}).
Due to prohibitively large size of MovieCLIP (in terms of hours) for human annotation, we leverage CLIP's zero-shot capabilities to tag the shots with scene labels.
\par 
Based on prompt engineering designs considered by GPT3 \cite{GPT3}, addition of contextual phrases like \textit{\textbf{``a type of pet"}} or \textit{\textbf{``a type of food"}} in the prompts provides additional information for CLIP for zero shot classification. In a similar vein, we consider the visual scene specific prompt: \textit{\textbf{``A photo of a \{label\}, a type of background location"}} to tag individual frames in video clips with labels from our scene taxonomy.
If a shot contains $T$ frames, we utilize CLIP's visual encoder to extract frame-wise visual embeddings $v_{t}(t=1,...,T)$. For each of the individual scene labels in our taxonomy, we utilize CLIP's text encoder to extract embeddings $e_{l} (l=1,2,...,L)$ for the background specific prompts 
We use the label-wise (prompt-specific) text and frame-wise visual embeddings to obtain a similarity score matrix $S$, whose entries $S_{lt}$ are computed as follows:
\begin{equation}\label{simmatrixscore}
    S_{lt}=\frac{e_{l}^{T}v_{t}}{\lVert e_{l}  \rVert_{2}  \lVert v_{t} \rVert_{2}}
\end{equation}
We compute an aggregate shot specific score for individual scene labels by temporal average pooling over the similarity matrix $S_{lt}$, since the visual content within a shot remains fairly unchanged. The computation of shot specific score called $\texttt{CLIPSceneScore}_{\texttt{l}}$ for $l$th visual scene label is shown in Eq.~\ref{score pooling}.
Overall workflow of the process is illustrated in Fig \ref{clip tagging}.
\begin{equation}
    \texttt{CLIPSceneScore}_{\texttt{l}}=\frac{\sum_{t=0}^{T}(S_{lt})}{T}
    \label{score pooling}
\end{equation}

\subsection{Analysis of the CLIP labeling}
\begin{figure}[h!]
\centering
\subfloat[]{%
\includegraphics[width=0.7\columnwidth]{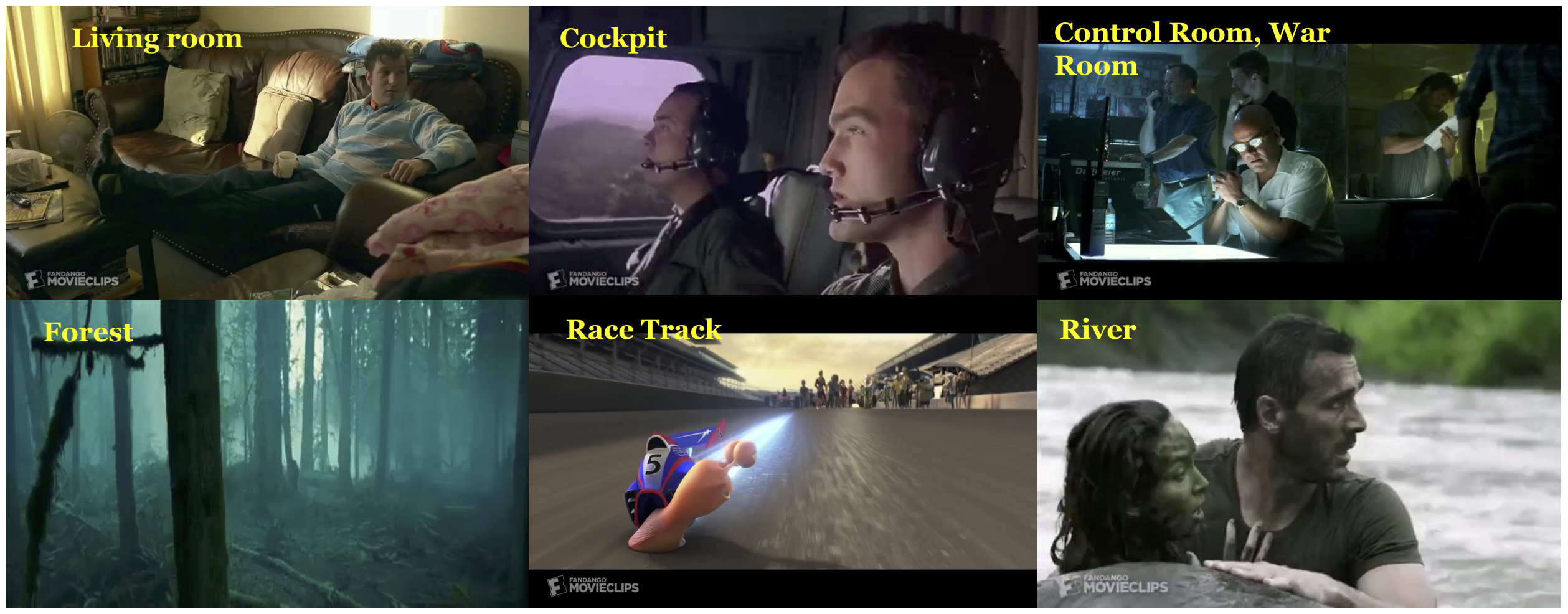}% 
\label{clip high conf}
}

\subfloat[]{%
\includegraphics[width=0.5\columnwidth]{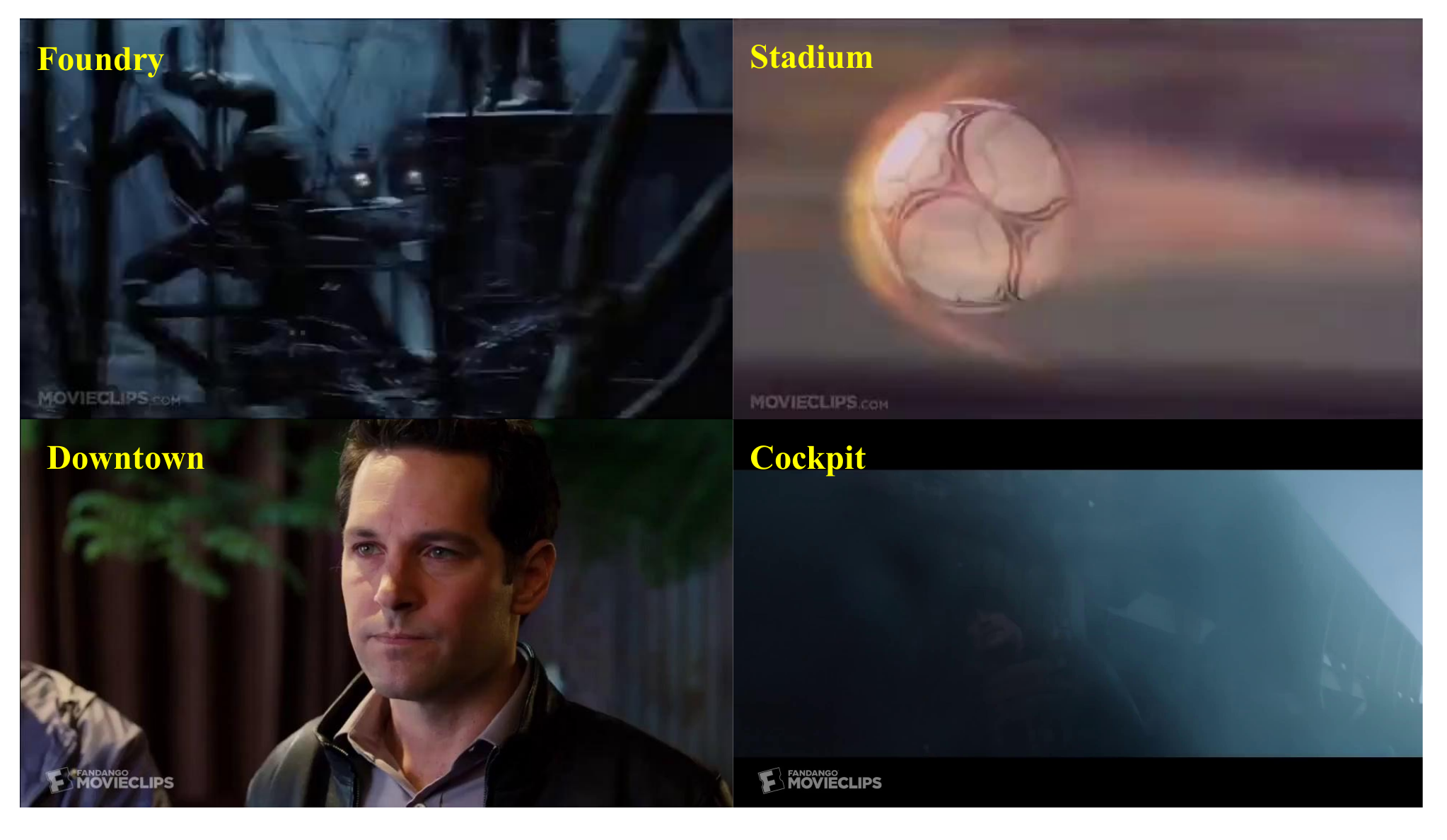}%
\label{clip low conf}
}
\caption{(a) Sample frames from the movie shots labeled by CLIP with high confidence (\texttt{CLIPSceneScore} $\geq$ 0.6 and Labels shown in yellow) (b) Sample frames from the movie shots tagged by CLIP with low confidence (Labels shown in yellow).}
\label{combined slulgline and image datasets}
\end{figure}
\begin{figure}[h!]
    \centering
    \includegraphics[width=0.8\columnwidth]{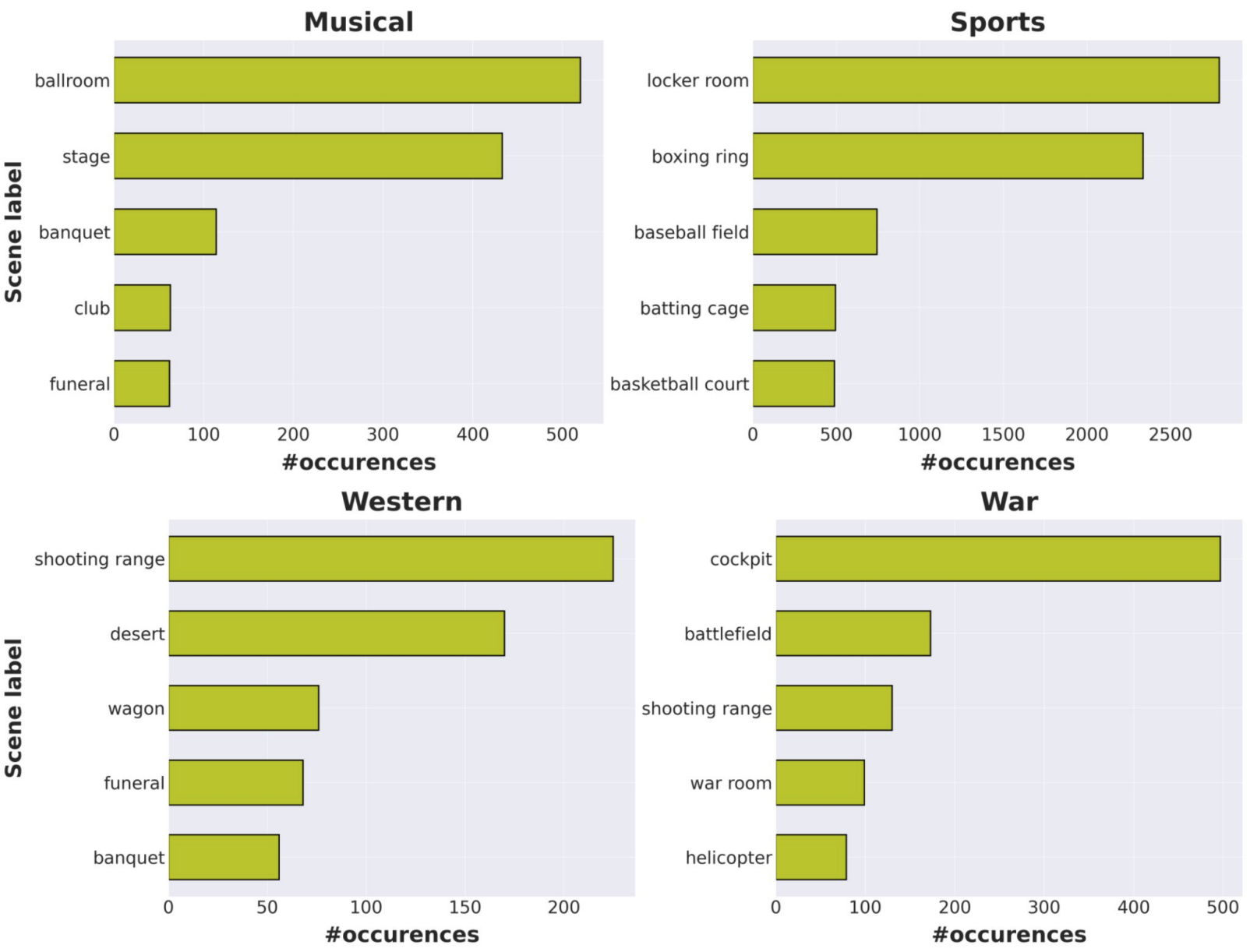}
    \caption{Genre wise distribution of different scene labels. For each genre, the top-5 scene labels are shown in terms of number of occurrences in the top-1 label provided by CLIP \cite{CLIP} for shots in MovieCLIP dataset. Threshold for confidence score of top-1 label = 0.4}
    \label{genre distribution}
\end{figure}
\begin{figure}
    \centering
    \includegraphics[width=0.8\columnwidth]{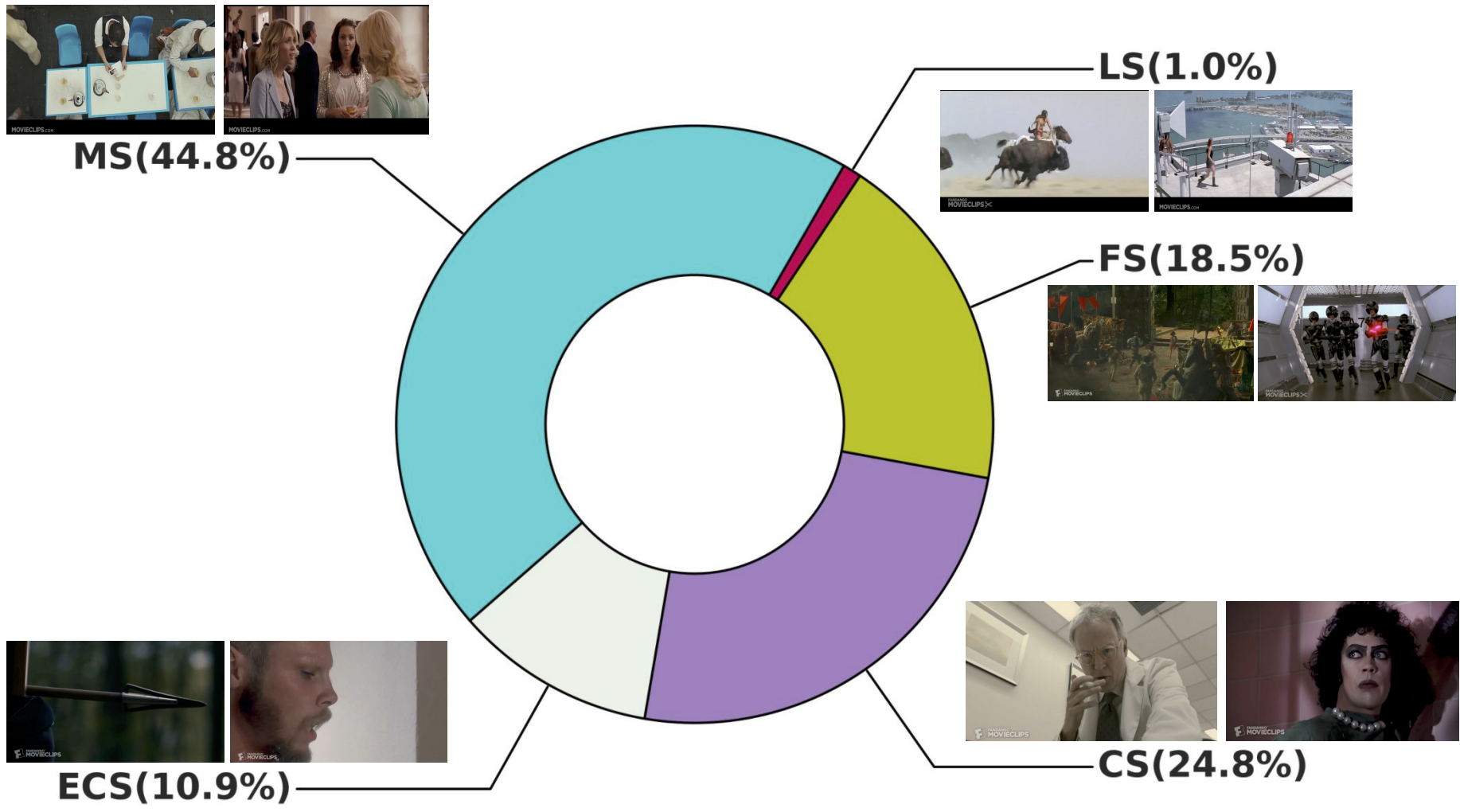}
    \caption{Distribution of shot scale predictions among the shots having no agreements between human annotators and CLIP's labeling scheme. \textbf{ECS}: Extreme Close-up shot, \textbf{CS}: Close-up shot, \textbf{MS}: Medium shot, \textbf{LS}: Long shot, \textbf{FS}: Full shot}
    \label{shot_type_distribution}
\end{figure}

\textbf{Qualitative analysis:}
%We qualitatively analyze CLIP's labeling performance on movie shots and diagnose failure modes where CLIP fails to associate visual scene labels with high confidence. 
As shown in Fig \ref{clip high conf}, CLIP performs well when distinctive elements of visual scene are present like background objects in indoor locations (living room) or appearance based cues (green color background for forests).
For example in  \ref{clip high conf}, the presence of airplane windows indicate that the associated visual scene label with the given movie shot is cockpit. However, for shots involving blurry motion or absence of background information due to close up of the involved characters, CLIP's confidence in associating visual scene labels is low (Fig \ref{clip low conf}).
\\
\textbf{Genre-wise association:} 
We consider those shots whose top-1 $\texttt{CLIPSceneScore}$ values are greater than or equal to 0.4 and show the top-5 scene labels in terms of occurrences for certain genres like \textit{western}, \textit{sport}, \textit{war} and \textit{musical}.  From Fig \ref{genre distribution}, we can see that relevant scenes are associated with genres through CLIP's labeling scheme. Some notable examples include \{shooting range, desert\} for \textit{western}, \{locker room, boxing ring\} for \textit{sport}, \{ballroom, stage\}  for \textit{musical} and \{cockpit, battlefield\} for \textit{war}.
\\
\textbf{Reliablity estimation:} In order to estimate reliability of the top-k labels provided by CLIP for movie shots, we conduct a verification task on Amazon Mechanical Turk. We provided a pool of annotators with a subset of 2393 movie shots from VidSitu \cite{Sadhu_2021_CVPR} dataset, along with top-5 scene labels. 
Out of the provided top-5 scene labels, the annotators were asked to select all the labels that apply for the given movie shot. We find 48.4\% and 80\% agreements among annotators in marking the top-1 and top-5 CLIP labels as relevant. After the human verification experiment is complete, we discard the shot samples with no agreements and obtain an evaluation data of 1883 shot samples.
\\
\textbf{Shot-type association:} We consider the shots discarded in the reliability estimation phase to analyze distribution with various shot scale types. Based on the shot scale labels available as part of MovieShots dataset \cite{rao2020unified}, we train a 2-layer LSTM (hidden dim= 512) \cite{lstm} network using frame wise features from a pretrained ViT-B/16 \cite{dosovitskiy2020vit} (extracted at 4 fps). The distribution of shot-wise predictions from the trained LSTM model is shown in Fig \ref{shot_type_distribution}. We can see that 80\% of shots having no agreements with human annotations belong to the shot categories having moderate (MS) to very high person closeup (ECS). 

% \begin{figure*}[h!]
%     \centering
%     \includegraphics[width=0.8\textwidth]{Figures/Genre_wise_plots.png}
%     \caption{Genre wise distribution of different scene labels. For each genre, the top-5 scene labels are shown in terms of number of occurrences in the top-1 label provided by CLIP \cite{CLIP} for shots in movieCLIP dataset. Threshold for confidence score of top-1 label = 0.4}
%     \label{genre distribution}
% \end{figure*}
\section{Experiments and Results}
\subsection{Experimental Setup}
For training and validation purposes, we retain those shot samples whose top-1 $\texttt{CLIPSceneScore}$ is greater than or equal to 0.4 (approx 75 percentile), resulting in a clean subset. After top-1 filtering, we also consider labels from top-k (k = 2 to 5) whose $\texttt{CLIPSceneScore}$ is greater than 0.1 to associate multiple labels per sample.
This results in a set of 107k samples with train, val and test split of 73.8k, 23.2k and 10.3k  having non-intersecting set of ids with the human-verified evaluation set. Approximately 38.4\% of the dataset is multi-label, covering 150 scene classes out of 179 in the curated scene taxonomy. 
All the related experiments were conducted using the Pytorch\cite{Pytorch} framework using 4 T4 NVIDIA GPUs. For training the respective models, we use binary cross entropy loss function. For evaluation, we use mean average precision (mAP) and Pearson correlation (averaged across samples) as metrics.
%For visual scene recognition, we use mean average precision (mAP) and Pearson correlation (averaged across samples) for evaluating the multi-label models. For downstream tasks, we report mean average precision (mAP) for the respective models.
\subsection{Visual scene recognition - Movies}
% We consider the task of multi label visual scene recognition from individual movie shots and explore modeling along three different aspects:\\
\textbf{Frame wise aggregation models:} For frame wise aggregation, we extract dense embeddings from individual shots at 4fps. We use two sets of embeddings: 512 dim embedding from Resnet18 \cite{He2015} pretrained on Places2 dataset and 768 dim embedding from ViT-B/16 \cite{dosovitskiy2020vit} model pretrained on Imagenet \cite{deng2009imagenet}.
Following feature extraction, we perform temporal aggregation using LSTM \cite{lstm} with 2 layers and hidden dimension of 512.
\\
\textbf{3D Convolutional network models:} 
We use I3D\cite{i3d}, R(2+1)D \cite{r2plus1d}, Slowfast \cite{feichtenhofer2019slowfast} as baseline 3D convolutional models in multi-label setup. I3D\cite{i3d} and Slowfast\cite{feichtenhofer2019slowfast} models have a Resnet50\cite{He2015} backbone whereas R(2+1)D \cite{r2plus1d} has a Resnet34\cite{He2015} backbone. All the models are initialized from Kinetics400 \cite{kinetics400} pretrained weights. For finetuning I3D\cite{i3d} and Slowfast\cite{feichtenhofer2019slowfast}, we use SGD with learning rates in $\{0.1,1e-3\}$ and weight decay of 1e-4. For R(2+1)D \cite{r2plus1d} we use Adam \cite{Kingma2015AdamAM} with learning rate 1e-4. Batch sizes for the models are varied between 16 and 32. \\
\textbf{Video Transformer models:}
For video transformer models, we consider the base TimeSformer model \cite{Bertasius2021IsSA} that considers 8 frames (224 x 224) as inputs. For finetuning TimeSformer \cite{Bertasius2021IsSA} model, we use SGD with learning rate 5e-3 and weight decay of 1e-4 and batch size of 8. For better speed-accuracy tradeoff we use the Video Swin Transformer model \cite{liu2021video}, \cite{liu2021Swin} called Swin-B with clip size of 32 frames (224 x 224) as inputs. For finetuning we use AdamW \cite{Loshchilov2019DecoupledWD} optimizer with learning rate 1e-4 and Cosine Annealing with batch size of 32. More details of the hyperparameter settings for above mentioned models are included in Supplementary.
Based on the results in Table \ref{visual movie models}, we can see that 2 layer LSTM models trained using features from Imagenet-21K pretrained ViT-B/16 perform better compared to features extracted using Resnet-18 model pretrained on Places2 dataset. This shows that features from Places2 pretrained models might not be optimal for scene recognition in the movie domain.
In terms of end-to-end models, video transformers including TimeSformer and Swin-B models outperform 3D convolutional models. Swin-B model performs better than other models by obtaining an average correlation of 0.497 and mean average precision of 44.4.
\vspace{-2 mm}
\begin{table}[h!]
\centering
\resizebox{\columnwidth}{!}{
\begin{tabular}{|cccc|}
\hline
\multicolumn{4}{|c|}{\textbf{Frame wise aggregation}}                                                                         \\ \hline
\multicolumn{1}{|c|}{Model}                & \multicolumn{1}{c|}{Features}         & \multicolumn{1}{c|}{mAP}   & Correlation \\ \hline
\multicolumn{1}{|l|}{LSTM (512, 2 layers)} & \multicolumn{1}{c|}{Places2 (4 fps)}  & \multicolumn{1}{c|}{24.15} & 0.29        \\ \hline
\multicolumn{1}{|l|}{LSTM (512, 2 layers)} & \multicolumn{1}{l|}{ViT-B/16 (4 fps)} & \multicolumn{1}{c|}{43.10} & 0.42        \\ \hline
\multicolumn{4}{|c|}{\textbf{3D convolutional networks}}                                                                      \\ \hline
\multicolumn{1}{|c|}{Model}                & \multicolumn{1}{c|}{Features}         & \multicolumn{1}{c|}{mAP}   & Correlation \\ \hline
\multicolumn{1}{|c|}{SlowFast (R50) \cite{feichtenhofer2019slowfast}}       & \multicolumn{1}{c|}{NA}               & \multicolumn{1}{c|}{25.80}      &   0.402         \\ \hline
\multicolumn{1}{|c|}{R(2+1)D (R34) \cite{r2plus1d} }        & \multicolumn{1}{c|}{NA}               & \multicolumn{1}{c|}{26.73} & 0.40        \\ \hline
\multicolumn{1}{|c|}{I3D (R50) \cite{i3d}}            & \multicolumn{1}{c|}{NA}               & \multicolumn{1}{l|}{13.33} & 0.26        \\ \hline
\multicolumn{4}{|c|}{\textbf{Video Transformers}}                                                                             \\ \hline
\multicolumn{1}{|c|}{TimeSformer \cite{Bertasius2021IsSA}}          & \multicolumn{1}{c|}{NA}               & \multicolumn{1}{l|}{36.87} & 0.46        \\ \hline
\multicolumn{1}{|c|}{\textbf{Swin-B} \cite{liu2021video}}             & \multicolumn{1}{c|}{\textbf{NA}}              & \multicolumn{1}{c|}{\textbf{44.4}}  & \textbf{0.497}       \\ \hline
\end{tabular}}
\vspace{1mm}
\caption{Mean average precision (mAP) and average Spearman correlation of different models on human verified evaluation set (N=1883 shots). NA: End-to-end models used instead of features. For 3D conv models, the backbone network is mentioned inside brackets. }
\label{visual movie models}
\end{table}
\subsection{Downstream tasks}
\subsubsection{Visual scene recognition - web videos:}
We also explore knowledge transfer from models finetuned on MovieCLIP by evaluating performance on downstream multi-label scene classification with HVU dataset \cite{diba_large_2020}. For training and evaluation, we use 251k and 16k videos with 248 scene labels. We extract 1024 dim features from the best performing Swin-B model in Table \ref{visual movie models}.
%, denoted by $S_{Movie\_scene}$.
%We extract 1024 dim features from the best performing Swin-B model in Table \ref{visual movie models} and Swin-B model pretrained on Kinetics400, denoted as $S_{Movie}$ and $S_{Base}$ respectively. 
We train 3 layer fully connected models on the respective features with the following configuration:\\
\textit{$M_{scene}$}: $\textbf{INP}[1024]\rightarrow{}\textbf{FC}[4096],\textbf{DO}(0.2)
\rightarrow{}\textbf{FC}[4096] \\ \rightarrow{}\textbf{FC}[248]$\\
From Table \ref{HVU}, we can see that ${M_{scene}}$ exhibits better performance when compared to existing end-to-end models trained on HVU.
\begin{table}[h!]
\centering
\begin{tabular}{|c|c|}
\hline
Model                                     & mAP \\ \hline
3D-ResNet \cite{diba_large_2020}                                &   50.6  \\ \hline 
3D-STCNet  \cite{diba_large_2020}                            &  51.9   \\ \hline
HATNet  \cite{diba_large_2020}                              &  55.8   \\ \hline
${M_{scene}}$ &  55.92   \\ \hline
\end{tabular}
\vspace{5mm}
\caption{Mean average precision of different models on HVU dataset for multi label scene classification (248 classes). Backbone for end to end models: 3D Resnet18.}
\label{HVU}
\end{table}
\vspace{- 7 mm}
\subsubsection{Multi label genre classification - Movie trailers:}

\begin{table*}[h!]
\centering
\resizebox{\textwidth}{!}{
\begin{tabular}{|c|c|c|c|c|c|c|c|c|c|c|c|c|c|c|}
\hline
\textbf{Model} & \textbf{Overall} & \textbf{Ac} & \textbf{Ani} & \textbf{Bio} & \textbf{Com} & \textbf{Cri} & \textbf{Drm} & \textbf{Fmy} & \textbf{Fntsy} & \textbf{Hrrr} & \textbf{Myst} & \textbf{Rom} & \textbf{ScF} & \textbf{Thrl} \\ \hline
$M_{trailer}$  & 56.14  &  62.97   &  86.51   &  14.4  &  80.77 &  49.58  &  79.58   &  74.55  &  49.59  &  50.62  &  26.83 &  45.05   &   47.99 & 61.36 \\ \hline
% FCensm $^{\ast}$         & 56.26            & 62.42       & 87.39        & 14.66        & 81.07        & 49.36        & 79.82        & 74.9         & 48.64          & 51.71         & 28.5          & 44.82        & 47.93        & 60.2          \\ \hline
C3D  \cite{DuTran}         & 53.4             & 63.8        & 91.3         & 16.2         & 82.3         & 45.1         & 71.6         & 65.3         & 54.8           & 50.8          & 28.2          & 38.3         & 21.8         & 64.8          \\ \hline
I3D  \cite{i3d}          & 38.8             & 37.2        & 51.8         & 9.2          & 72.6         & 33.9         & 67.6         & 43.6         & 39             & 22.8          & 21.3          & 34.3         & 22.6         & 48.3          \\ \hline
LSTM \cite{2019Moviescope}           & 48.4             & 47.5        & 86.8         & 12           & 79.2         & 33           & 72           & 64.5         & 54.4           & 22.7          & 24.7          & 40.4         & 36.5         & 54.8          \\ \hline
Bi-LSTM  \cite{2019Moviescope}       & 47.4             & 49.9        & 86.3         & 8.2          & 77.6         & 29.9         & 70.8         & 65.4         & 55.3           & 22.3          & 21.7          & 41.6         & 35.9         & 51.2          \\ \hline
fstVid \cite{2019Moviescope}      & 56.5             & 61.4        & 94.8         & 23.9         & 81.5         & 41.7         & 77           & 67           & 62.6           & 36.1          & 30.4          & 48.4         & 48.2         & 62            \\ \hline
fstTConv \cite{2019Moviescope} & 58.9             & 64.7        & 95.7         & 21.2         & 83.5         & 49.1         & 78.9         & 68.6         & 68.9           & 42.7          & 29.2          & 46.8         & 51           & 64.8          \\ \hline
\end{tabular}
}
\vspace{5mm}
\caption{Mean average precision of different models for multi-label genre classification (13 class) on Moviescope dataset. Except $M_{trailer}$ comparison results are reported from \cite{2019Moviescope}. Abbreviations: \textbf{Ac:} Action, \textbf{Ani:} Animation, \textbf{Bio:} Biography, \textbf{Com}: Comedy, \textbf{Cri}: Crime, \textbf{Drm}: Drama, \textbf{Fmy}: Family, \textbf{Fntsy}: Fantasy, \textbf{Hrrr}: Horror, \textbf{Myst}: Mystery, \textbf{Rom}: Romantic, \textbf{ScF}: SciFi, \textbf{Thrl}: Thriller, \textbf{fstVid}: fastVideo, \textbf{fstTConv}: fastVideo + Temporal Conv. }
\label{genre}
\end{table*}
As an additional downstream task, we consider multi-label genre classification of movie trailers in the Moviescope dataset \cite{2019Moviescope}. Out of the original set of 4927 trailers, we could access 3900 videos from YouTube. Based on the provided splits, we use 2948, 410 and 542 videos for training, validation and testing purposes, respectively. We use the 1024 dim features extracted from the best performing Swin-B model in Table \ref{visual movie models}.
%, denoted by $S_{Movie\_genre}$. 
We train 3 layer fully connected models on the respective features with the following configuration:
\textit{$M_{trailer}$}: $\textbf{INP}[1024]\rightarrow{} \textbf{FC}[512],\textbf{DO}(0.2)\rightarrow{}\textbf{FC}[512]\rightarrow{}\textbf{FC}[13]$\\
Even when the number of trailer videos used is a subset of the original split, $M_{trailer}$ exhibits similar genre wise trends as other models. From Table \ref{genre}, we can see that $M_{trailer}$ shows better performance in Animation and Comedy as compared to genres like Biography and Mystery. When compared with fstTConv in Table \ref{genre}, our fully connected model $M_{trailer}$ performs slightly worse due to non-availability of entire training data. 
%and the features from Swin-B are extracted for the entire trailer (around 2 mins in length) instead of aggregating shot specific features. 
\subsubsection{Impact of MovieCLIP pretraining:}
\begin{table}[h!]
\centering
\resizebox{0.9\columnwidth}{!}{
\begin{tabular}{|cc|cc|}
\hline
\multicolumn{2}{|c|}{\textbf{HVU}}         & \multicolumn{2}{c|}{\textbf{Moviescope}}    \\ \hline
\multicolumn{1}{|c|}{Model}    & mAP & \multicolumn{1}{c|}{Model}      & mAP \\ \hline
\multicolumn{1}{|c|}{$M_{scene}$} &   55.92  & \multicolumn{1}{c|}{$M_{trailer}$} &  56.14   \\ \hline
\multicolumn{1}{|c|}{$M_{scene}(Kin)$}  &  56.05   & \multicolumn{1}{c|}{$M_{trailer}(Kin)$}  & 53.29    \\ \hline
\multicolumn{1}{|c|}{Late Fusion}             &  57.73   & \multicolumn{1}{c|}{Late Fusion}            &   56.29   \\ \hline
\end{tabular}
}
\vspace{3mm}
\caption{Impact of MovieCLIP pretrained features vs Kinetics pretrained features for $M_{scene}$ (HVU) and $M_{trailer}$ (Moviescope). Results reported are mean average precision (mAP) values. $Model(Kin)$: Model with Kinetics400 pretrained features, where $Model \in \{M_{scene},M_{trailer}\}$ }
\label{pretrain table}.
\end{table}
We consider the impact of MovieCLIP based pretraining by fixing the fully connected architectures $M_{scene}$, $M_{trailer}$ and varying the input features. In the without MovieCLIP ($Kin$) pretraining setting, we extract 1024 dim features from Swin-B model pretrained on Kinetics400 for HVU and Moviescope datasets. 

From Table \ref{pretrain table}, we can see that the performance of $M_{scene}$ with MovieCLIP pretrained features is comparable to $M_{scene}(Kin)$, even when the domain of Kinetics400 \cite{kinetics400} is matched to HVU. 
% From Table \ref{pretrain table}, we can see that the performance obtained using $S_{Movie\_scene}$ features for HVU is comparable to $S_{Base\_scene}$, even when the domain of Kinetics400 \cite{kinetics400} is matched to HVU. 
Further, late fusion of prediction logits of $M_{scene}$ and  $M_{scene}(Kin)$ with equal weights improves the mAP to 57.73 for HVU, thus indicating capture of complementary information, when trained with movie data. We include a class wise analysis of HVU dataset in Supplementary (Figure 11) to showcase the classes where MovieCLIP pretrained features improve upon Kinetics400 pretrained features. In case of Moviescope, $M_{trailer}$ 
results in improved performance (\textbf{56.14}) as compared to $M_{trailer}(Kin)$ (\textbf{53.29}), due to domain similarity with MovieCLIP dataset. 

\section{Ethical implications}
Visual scene recognition capabilities can help in uncovering biases associated with the portrayal of under-represented and marginalized characters in various settings. For example, women are portrayed more in indoor scenes like kitchen, living room, hospital as compared to scenes like factory, laboratory or battlefield. Further, characters from marginalized demographic groups are often depicted in the background w.r.t common visual scenes, thus having considerable less share of speaking time. Apart from portrayal of characters, the usage of large scale pretrained models like CLIP \cite{CLIP} can help diagnose the inherent biases associated with its predictions since it is trained on free-form data curated from web. The proposed scheme of utilizing CLIP for weakly tagging datasets can reduce the costs associated with large-scale human expert driven annotation process.

\section{Conclusion}
In this work, we introduce a rich movie-centric taxonomy of visual scene labels, automatically curated from movie scripts and HVU \cite{diba_large_2020}, with minimal human-in-the loop intervention. Further, we utilize CLIP's \cite{CLIP} zero-shot capabilities to weakly label movie shots based on our curated taxonomy in a scalable manner. %The taxonomy curation process and subsequent CLIP tagging is scalable since the entire pipeline can be repeated for an updated set of movie scripts and auxiliary sources. 
We develop baseline end-to-end models on the weakly labelled dataset called MovieCLIP and evaluate on an independent human-verified dataset with scene labels. We explore the utility of MovieCLIP dataset as a pretraining source by evaluating on two downstream tasks of multi-label scene and genre classification of web videos \cite{diba_large_2020} and movie trailers \cite{2019Moviescope}. Future directions include modeling of temporal transitions of visual scenes across shots, multimodal association between audio events and visual scenes and multi-task modeling of visual scenes and related attributes like actions, time of day and settings (interior and exterior).

\section{Acknowledgements}
We would like to thank Boqing Gong for his feedback on the paper and help with experiment design. 
This work was supported by Google.

{\small
\bibliographystyle{ieee_fullname}
\bibliography{egbib}

\begin{thebibliography}{10}\itemsep=-1pt

\bibitem{bain2020condensed}
Max Bain, Arsha Nagrani, Andrew Brown, and Andrew Zisserman.
\newblock Condensed movies: Story based retrieval with contextual embeddings,
  2020.

\bibitem{BarsoumICMI2016}
Emad Barsoum, Cha Zhang, Cristian Canton~Ferrer, and Zhengyou Zhang.
\newblock Training deep networks for facial expression recognition with
  crowd-sourced label distribution.
\newblock In {\em ACM International Conference on Multimodal Interaction
  (ICMI)}, 2016.

\bibitem{Bertasius2021IsSA}
Gedas Bertasius, Heng Wang, and Lorenzo Torresani.
\newblock Is space-time attention all you need for video understanding?
\newblock {\em ArXiv}, abs/2102.05095, 2021.

\bibitem{bordwellthomson}
David Bordwell and Kirstin Thomson.
\newblock {\em Film art: An introduction}.
\newblock McGraw Hill, 2001, 2017.

\bibitem{bossard14}
Lukas Bossard, Matthieu Guillaumin, and Luc Van~Gool.
\newblock Food-101 -- mining discriminative components with random forests.
\newblock In {\em European Conference on Computer Vision}, 2014.

\bibitem{GPT3}
Tom Brown, Benjamin Mann, Nick Ryder, Melanie Subbiah, Jared~D Kaplan, Prafulla
  Dhariwal, Arvind Neelakantan, et~al.
\newblock Language models are few-shot learners.
\newblock In {\em Advances in Neural Information Processing Systems},
  volume~33, pages 1877--1901. Curran Associates, Inc., 2020.

\bibitem{Kinetics700}
Jo{\~{a}}o Carreira, Eric Noland, Chloe Hillier, and Andrew Zisserman.
\newblock A short note on the kinetics-700 human action dataset.
\newblock {\em CoRR}, abs/1907.06987, 2019.

\bibitem{i3d}
J. Carreira and Andrew Zisserman.
\newblock Quo vadis, action recognition? a new model and the kinetics dataset.
\newblock pages 4724--4733, 07 2017.

\bibitem{2019Moviescope}
Paola Cascante-Bonilla, Kalpathy Sitaraman, Mengjia Luo, and Vicente Ordonez.
\newblock Moviescope: Large-scale analysis of movies using multiple modalities.
\newblock {\em ArXiv}, abs/1908.03180, 2019.

\bibitem{10.1007/978-3-540-88693-8_12}
Timothee Cour, Chris Jordan, Eleni Miltsakaki, and Ben Taskar.
\newblock Movie/script: Alignment and parsing of video and text transcription.
\newblock In David Forsyth, Philip Torr, and Andrew Zisserman, editors, {\em
  Computer Vision -- ECCV 2008}, pages 158--171, Berlin, Heidelberg, 2008.
  Springer Berlin Heidelberg.

\bibitem{deng2009imagenet}
Jia Deng, Wei Dong, Richard Socher, Li-Jia Li, Kai Li, and Li Fei-Fei.
\newblock Imagenet: A large-scale hierarchical image database.
\newblock In {\em 2009 IEEE conference on computer vision and pattern
  recognition}, pages 248--255. Ieee, 2009.

\bibitem{desai2021virtex}
Karan Desai and Justin Johnson.
\newblock {VirTex: Learning Visual Representations from Textual Annotations}.
\newblock In {\em CVPR}, 2021.

\bibitem{diba_large_2020}
Ali Diba, Mohsen Fayyaz, Vivek Sharma, Manohar Paluri, Jürgen Gall, Rainer
  Stiefelhagen, and Luc Van~Gool.
\newblock Large {Scale} {Holistic} {Video} {Understanding}.
\newblock In Andrea Vedaldi, Horst Bischof, Thomas Brox, and Jan-Michael Frahm,
  editors, {\em Computer {Vision} – {ECCV} 2020}, pages 593--610, Cham, 2020.
  Springer International Publishing.

\bibitem{dosovitskiy2020vit}
Alexey Dosovitskiy, Lucas Beyer, Alexander Kolesnikov, Dirk Weissenborn,
  Xiaohua Zhai, Thomas Unterthiner, Mostafa Dehghani, Matthias Minderer, Georg
  Heigold, Sylvain Gelly, Jakob Uszkoreit, and Neil Houlsby.
\newblock An image is worth 16x16 words: Transformers for image recognition at
  scale.
\newblock {\em ICLR}, 2021.

\bibitem{Everingham2006HelloMN}
Mark Everingham, Josef Sivic, and Andrew Zisserman.
\newblock Hello! my name is... buffy'' -- automatic naming of characters in tv
  video.
\newblock In {\em BMVC}, 2006.

\bibitem{caba2015activitynet}
Bernard~Ghanem Fabian Caba~Heilbron, Victor~Escorcia and Juan~Carlos Niebles.
\newblock Activitynet: A large-scale video benchmark for human activity
  understanding.
\newblock In {\em Proceedings of the IEEE Conference on Computer Vision and
  Pattern Recognition}, pages 961--970, 2015.

\bibitem{FeiFei2004LearningGV}
Li Fei-Fei, Rob Fergus, and Pietro Perona.
\newblock Learning generative visual models from few training examples: An
  incremental bayesian approach tested on 101 object categories.
\newblock {\em Computer Vision and Pattern Recognition Workshop}, 2004.

\bibitem{BayesianFeiFeiLi}
L. Fei-Fei and P. Perona.
\newblock A bayesian hierarchical model for learning natural scene categories.
\newblock In {\em 2005 IEEE Computer Society Conference on Computer Vision and
  Pattern Recognition (CVPR'05)}, volume~2, pages 524--531 vol. 2, 2005.

\bibitem{feichtenhofer2019slowfast}
Christoph Feichtenhofer, Haoqi Fan, Jitendra Malik, and Kaiming He.
\newblock Slowfast networks for video recognition.
\newblock In {\em Proceedings of the IEEE international conference on computer
  vision}, pages 6202--6211, 2019.

\bibitem{brendanfrey}
Brendan~J. Frey and Delbert Dueck.
\newblock Clustering by passing messages between data points.
\newblock {\em Science}, 315(5814):972--976, 2007.

\bibitem{Something-SomethingV2}
Raghav Goyal, Samira~Ebrahimi Kahou, Vincent Michalski, Joanna Materzynska,
  Susanne Westphal, et~al.
\newblock The "something something" video database for learning and evaluating
  visual common sense.
\newblock {\em CoRR}, abs/1706.04261, 2017.

\bibitem{gu2018ava}
Chunhui Gu, Chen Sun, David~A Ross, Carl Vondrick, Caroline Pantofaru, et~al.
\newblock Ava: A video dataset of spatio-temporally localized atomic visual
  actions.
\newblock In {\em Proceedings of the IEEE Conference on Computer Vision and
  Pattern Recognition}, pages 6047--6056, 2018.

\bibitem{Gu2021OpenvocabularyOD}
Xiuye Gu, Tsung-Yi Lin, Weicheng Kuo, and Yin Cui.
\newblock Open-vocabulary object detection via vision and language knowledge
  distillation.
\newblock 2021.

\bibitem{He2015}
Kaiming He, Xiangyu Zhang, Shaoqing Ren, and Jian Sun.
\newblock Deep residual learning for image recognition.
\newblock {\em arXiv preprint arXiv:1512.03385}, 2015.

\bibitem{SBD}
Daniel Helm and Martin Kampel.
\newblock Shot boundary detection for automatic video analysis of historical
  films.
\newblock In Marco Cristani, Andrea Prati, Oswald Lanz, Stefano Messelodi, and
  Nicu Sebe, editors, {\em New Trends in Image Analysis and Processing -- ICIAP
  2019}, pages 137--147, Cham, 2019. Springer International Publishing.

\bibitem{lstm}
Sepp Hochreiter and J{\"u}rgen Schmidhuber.
\newblock Long short-term memory.
\newblock {\em Neural computation}, 9(8):1735--1780, 1997.

\bibitem{huang2020movienet}
Qingqiu Huang, Yu Xiong, Anyi Rao, Jiaze Wang, and Dahua Lin.
\newblock Movienet: A holistic dataset for movie understanding.
\newblock In {\em The European Conference on Computer Vision (ECCV)}, 2020.

\bibitem{kinetics400}
Will Kay, Jo{\~{a}}o Carreira, Karen Simonyan, Brian Zhang, Chloe Hillier,
  Sudheendra Vijayanarasimhan, Fabio Viola, Tim Green, Trevor Back, Paul
  Natsev, Mustafa Suleyman, and Andrew Zisserman.
\newblock The kinetics human action video dataset.
\newblock {\em CoRR}, abs/1705.06950, 2017.

\bibitem{Kingma2015AdamAM}
Diederik~P. Kingma and Jimmy Ba.
\newblock Adam: A method for stochastic optimization.
\newblock {\em CoRR}, abs/1412.6980, 2015.

\bibitem{KrauseStarkDengFei-Fei_3DRR2013}
Jonathan Krause, Michael Stark, Jia Deng, and Li Fei-Fei.
\newblock 3d object representations for fine-grained categorization.
\newblock In {\em 4th International IEEE Workshop on 3D Representation and
  Recognition (3dRR-13)}, Sydney, Australia, 2013.

\bibitem{Laptevactionscvpr}
Ivan Laptev, Marcin Marszalek, Cordelia Schmid, and Benjamin Rozenfeld.
\newblock Learning realistic human actions from movies.
\newblock In {\em 2008 IEEE Conference on Computer Vision and Pattern
  Recognition}, pages 1--8, 2008.

\bibitem{Li2022LanguagedrivenSS}
Boyi Li, Kilian~Q. Weinberger, Serge~J. Belongie, Vladlen Koltun, and Ren{\'e}
  Ranftl.
\newblock Language-driven semantic segmentation.
\newblock {\em ArXiv}, abs/2201.03546, 2022.

\bibitem{liu2021Swin}
Ze Liu, Yutong Lin, Yue Cao, Han Hu, Yixuan Wei, Zheng Zhang, Stephen Lin, and
  Baining Guo.
\newblock Swin transformer: Hierarchical vision transformer using shifted
  windows.
\newblock {\em arXiv preprint arXiv:2103.14030}, 2021.

\bibitem{liu2021video}
Ze Liu, Jia Ning, Yue Cao, Yixuan Wei, Zheng Zhang, Stephen Lin, and Han Hu.
\newblock Video swin transformer.
\newblock {\em arXiv preprint arXiv:2106.13230}, 2021.

\bibitem{Loshchilov2019DecoupledWD}
Ilya Loshchilov and Frank Hutter.
\newblock Decoupled weight decay regularization.
\newblock In {\em ICLR}, 2019.

\bibitem{marszalek09}
Marcin Marsza{\l}ek, Ivan Laptev, and Cordelia Schmid.
\newblock Actions in context.
\newblock In {\em IEEE Conference on Computer Vision \& Pattern Recognition},
  2009.

\bibitem{statisticsmovie}
José~Gabriel Navarro.
\newblock Film industry in the u.s. - statistics \& facts.
\newblock \url{https://www.statista.com/topics/964/film/#dossierKeyfigures},
  2021.

\bibitem{parkhi12a}
O.~M. Parkhi, A. Vedaldi, A. Zisserman, and C.~V. Jawahar.
\newblock Cats and dogs.
\newblock In {\em IEEE Conference on Computer Vision and Pattern Recognition},
  2012.

\bibitem{Pytorch}
Adam Paszke, Sam Gross, Francisco Massa, Adam Lerer, James Bradbury, Gregory
  Chanan, Trevor Killeen, et~al.
\newblock Pytorch: An imperative style, high-performance deep learning library.
\newblock In H. Wallach, H. Larochelle, A. Beygelzimer, F. d\textquotesingle
  Alch\'{e}-Buc, E. Fox, and R. Garnett, editors, {\em Advances in Neural
  Information Processing Systems 32}, pages 8024--8035. Curran Associates,
  Inc., 2019.

\bibitem{IndoorScenes}
Ariadna Quattoni and Antonio Torralba.
\newblock Recognizing indoor scenes.
\newblock In {\em 2009 IEEE Conference on Computer Vision and Pattern
  Recognition}, pages 413--420, 2009.

\bibitem{CLIP}
Alec Radford, Jong~Wook Kim, Chris Hallacy, Aditya Ramesh, Gabriel Goh, et~al.
\newblock Learning transferable visual models from natural language
  supervision.
\newblock {\em CoRR}, abs/2103.00020, 2021.

\bibitem{rao2020unified}
Anyi Rao, Jiaze Wang, Linning Xu, Xuekun Jiang, Qingqiu Huang, Bolei Zhou, and
  Dahua Lin.
\newblock A unified framework for shot type classification based on subject
  centric lens.
\newblock In {\em The European Conference on Computer Vision (ECCV)}, 2020.

\bibitem{SOA}
Jamie Ray, Heng Wang, Du Tran, Yufei Wang, Matt Feiszli, Lorenzo Torresani, and
  Manohar Paluri.
\newblock Scenes-objects-actions: A multi-task, multi-label video dataset.
\newblock In {\em Computer Vision -- ECCV 2018}, pages 660--676, Cham, 2018.
  Springer International Publishing.

\bibitem{reimers-2019-sentence-bert}
Nils Reimers and Iryna Gurevych.
\newblock Sentence-bert: Sentence embeddings using siamese bert-networks.
\newblock In {\em Proceedings of the 2019 Conference on Empirical Methods in
  Natural Language Processing}. Association for Computational Linguistics, 11
  2019.

\bibitem{Rohrbach2015ADF}
Anna Rohrbach, Marcus Rohrbach, Niket Tandon, and Bernt Schiele.
\newblock A dataset for movie description.
\newblock {\em 2015 IEEE Conference on Computer Vision and Pattern Recognition
  (CVPR)}, pages 3202--3212, 2015.

\bibitem{Sadhu_2021_CVPR}
Arka Sadhu, Tanmay Gupta, Mark Yatskar, Ram Nevatia, and Aniruddha Kembhavi.
\newblock Visual semantic role labeling for video understanding.
\newblock In {\em The IEEE Conference on Computer Vision and Pattern
  Recognition (CVPR)}, June 2021.

\bibitem{sariyildiz2020icmlm}
Mert~Bulent Sariyildiz, Julien Perez, and Diane Larlus.
\newblock Learning visual representations with caption annotations.
\newblock In {\em European Conference on Computer Vision (ECCV)}, 2020.

\bibitem{Shen2021HowMC}
Sheng Shen, Liunian~Harold Li, Hao Tan, Mohit Bansal, Anna Rohrbach, Kai-Wei
  Chang, Zhewei Yao, and Kurt Keutzer.
\newblock How much can clip benefit vision-and-language tasks?
\newblock {\em ArXiv}, abs/2107.06383, 2021.

\bibitem{CMI}
Krishna Somandepalli, Tanaya Guha, Victor~R. Martinez, Naveen Kumar, Hartwig
  Adam, and Shrikanth Narayanan.
\newblock Computational media intelligence: Human-centered machine analysis of
  media.
\newblock {\em Proceedings of the IEEE}, 109(5):891--910, 2021.

\bibitem{ucf101}
Khurram Soomro, Amir~Roshan Zamir, and Mubarak Shah.
\newblock {UCF101:} {A} dataset of 101 human actions classes from videos in the
  wild.
\newblock {\em CoRR}, abs/1212.0402, 2012.

\bibitem{DuTran}
Du Tran, Lubomir Bourdev, Rob Fergus, Lorenzo Torresani, and Manohar Paluri.
\newblock Learning spatiotemporal features with 3d convolutional networks.
\newblock Dec. 2014.

\bibitem{r2plus1d}
Du Tran, Heng Wang, Lorenzo Torresani, Jamie Ray, Yann LeCun, and Manohar
  Paluri.
\newblock A closer look at spatiotemporal convolutions for action recognition.
\newblock In {\em Proceedings of the IEEE conference on Computer Vision and
  Pattern Recognition}, pages 6450--6459, 2018.

\bibitem{transformers}
Ashish Vaswani, Noam Shazeer, Niki Parmar, et~al.
\newblock Attention is all you need.
\newblock In I. Guyon, U.~V. Luxburg, S. Bengio, H. Wallach, R. Fergus, S.
  Vishwanathan, and R. Garnett, editors, {\em Advances in Neural Information
  Processing Systems}, volume~30. Curran Associates, Inc., 2017.

\bibitem{moviegraphs}
Paul Vicol, Makarand Tapaswi, Lluis Castrejon, and Sanja Fidler.
\newblock Moviegraphs: Towards understanding human-centric situations from
  videos.
\newblock In {\em {IEEE Conference on Computer Vision and Pattern Recognition
  (CVPR)}}, 2018.

\bibitem{lvu2021}
Chao-Yuan Wu and Philipp Kr\"{a}henb\"{u}hl.
\newblock {Towards Long-Form Video Understanding}.
\newblock In {\em {CVPR}}, 2021.

\bibitem{xiao_sun_2016}
Jianxiong Xiao, Krista~A. Ehinger, James Hays, Antonio Torralba, and Aude
  Oliva.
\newblock {SUN} {Database}: {Exploring} a {Large} {Collection} of {Scene}
  {Categories}.
\newblock {\em International Journal of Computer Vision}, 119(1):3--22, Aug.
  2016.

\bibitem{zhou2017places}
Bolei Zhou, Agata Lapedriza, Aditya Khosla, Aude Oliva, and Antonio Torralba.
\newblock Places: A 10 million image database for scene recognition.
\newblock {\em IEEE Transactions on Pattern Analysis and Machine Intelligence},
  2017.

\end{thebibliography}
}

\end{document}